\documentclass[conference]{IEEEtran}
\IEEEoverridecommandlockouts

\usepackage{spconf}
\usepackage[utf8]{inputenc}
\usepackage{url}
\usepackage{cite}
\usepackage{amsmath,amssymb,amsfonts}
\usepackage{algorithmic}
\usepackage{graphicx}
\usepackage{textcomp}
\usepackage{xcolor}
\usepackage{spconf,amsmath,graphicx}
\usepackage{multirow}
\usepackage{subfig}
\usepackage{url}
\def\BibTeX{{\rm B\kern-.05em{\sc i\kern-.025em b}\kern-.08em
    T\kern-.1667em\lower.7ex\hbox{E}\kern-.125emX}}

\makeatletter
\newcommand{\linebreakand}{%
    \end{@IEEEauthorhalign}
    \hfill\mbox{}\par
    \mbox{}\hfill\begin{@IEEEauthorhalign}
}
\makeatother
    
\begin{document}

\title{Can Large Language Models Grasp Event Signals? Exploring Pure Zero-Shot Event-based Recognition\\
{\footnotesize \textsuperscript{}}
}

\author{\IEEEauthorblockN{1\textsuperscript{st} Zongyou Yu}
\IEEEauthorblockA{\textit{School of Computer and Artificial Intelligence} \\
\textit{Beijing Technology and Business University}\\
Beijing, China \\
zongyou.yu@st.btbu.edu.cn}
\and
\IEEEauthorblockN{2\textsuperscript{nd} Qiang Qu}
\IEEEauthorblockA{\textit{School of Computer Science} \\
\textit{The University of Sydney}\\
Sydney, Australia \\
vincent.qu@sydney.edu.au}
\and
\IEEEauthorblockN{3\textsuperscript{rd} Xiaoming Chen}
\IEEEauthorblockA{\textit{School of Computer and Artificial Intelligence} \\
\textit{Beijing Technology and Business University}\\
Beijing, China \\
xiaoming.chen@btbu.edu.cn}
\linebreakand
\IEEEauthorblockN{4\textsuperscript{th} Chen Wang}
\IEEEauthorblockA{\textit{School of Computer and Artificial Intelligence} \\
\textit{Beijing Technology and Business University}\\
Beijing, China \\
wangc@btbu.edu.cn}
}

\maketitle

\begin{abstract}
Recent advancements in event-based zero-shot object recognition have demonstrated promising results. However, these methods heavily depend on extensive training and are inherently constrained by the characteristics of CLIP. To the best of our knowledge, this research is the first study to explore the understanding capabilities of large language models (LLMs) for event-based visual content. We demonstrate that LLMs can achieve event-based object recognition without additional training or fine-tuning in conjunction with CLIP, effectively enabling pure zero-shot event-based recognition. Particularly, we evaluate the ability of GPT-4o / 4turbo and two other open-source LLMs to directly recognize event-based visual content. Extensive experiments are conducted across three benchmark datasets, systematically assessing the recognition accuracy of these models. The results show that LLMs, especially when enhanced with well-designed prompts, significantly improve event-based zero-shot recognition performance. Notably, GPT-4o outperforms the compared models and exceeds the recognition accuracy of state-of-the-art event-based zero-shot methods on N-ImageNet by five orders of magnitude. The implementation of this paper is available at \url{https://github.com/ChrisYu-Zz/Pure-event-based-recognition-based-LLM}.
\end{abstract}

\begin{IEEEkeywords}
large language model, event camera, object recognition.
\end{IEEEkeywords}

\section{Introduction}
\label{sec:intro}

\begin{figure}[t]
  \centering
  \includegraphics[width=0.45\textwidth]{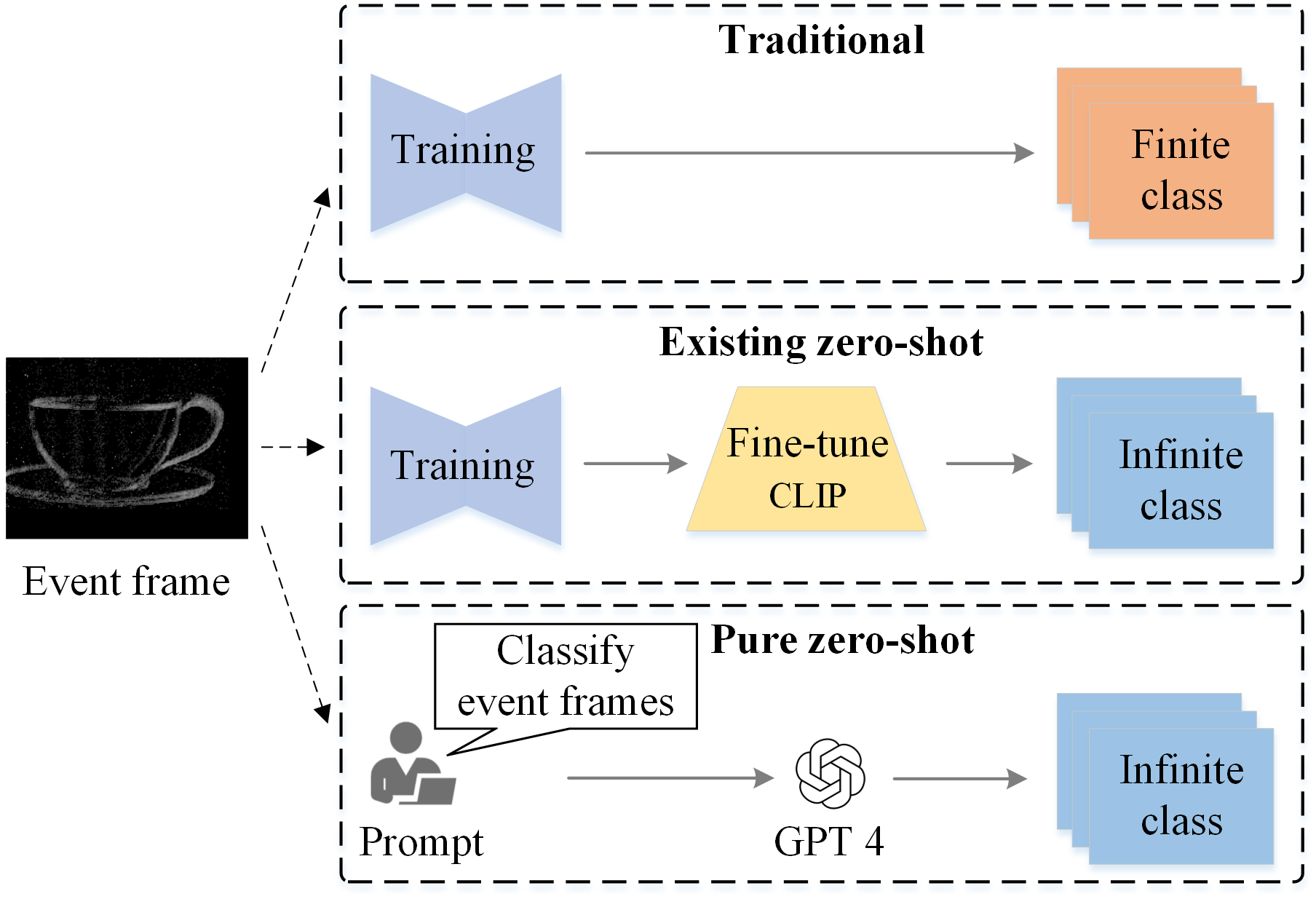}
  \caption{\textbf{Comparison of methods for event-based object recognition.} Unlike existing zero-shot approaches, our pure zero-shot recognition using LLMs eliminates the need for additional training, such as event encoder training or CLIP fine-tuning, while addressing the inherent limitations of CLIP.}
  \label{fig:teaser}
\end{figure}

Event-based cameras~\cite{b1}, due to their high temporal resolution, high dynamic range, and low power consumption, have garnered widespread attention in fields such as computer vision \cite{b2,b3}, achieving promising results in tasks like object recognition \cite{b4,b5}. Event-based object recognition methods can be classified into traditional and zero-shot approaches. As illustrated in Figure \ref{fig:teaser}, traditional methods require extensive training and are limited in the categories they can recognize due to the constraints of convolutional neural networks \cite{b6}. To address these limitations, researchers have proposed zero-shot methods that leverage large-scale pre-trained models \cite{b7}. Among these approaches, CLIP \cite{b8}, a groundbreaking model that leverages a contrastive learning strategy, aligns visual and textual representations within a shared embedding space. This alignment is particularly effective for zero-shot recognition tasks, where the model can associate images with textual descriptions of unseen categories. However, due to the domain gap between events and the associated textual labels, existing zero-shot methods are inherently constrained by the characteristics of CLIP. Additionally, existing zero-shot methods rely on an additional event encoder to process raw event data before classification, further complicating the recognition process.

Recent research has extensively explored the advanced capabilities of LLMs in visual tasks \cite{b9}. LLMs possess vast pre-trained knowledge, enabling them to excel in a variety of visual tasks, including event-based zero-shot scene understanding. While existing methods for event-based zero-shot object recognition show reasonable performance, they typically rely on some degree of training, rather than focusing solely on pure zero-shot recognition using event data. To address this problem, we propose utilizing LLMs to address the challenges inherent in event-based zero-shot object recognition. LLMs can process text and image inputs simultaneously, enabling the direct conversion of raw event streams into frames for model input, thus removing the need for a separate event encoder. Furthermore, the vast training datasets and embedded knowledge in LLMs enhance zero-shot recognition, overcoming the constraints of models like CLIP and improving recognition accuracy. In light of these advantages, this study systematically evaluates the efficacy of LLMs in achieving pure event-based zero-shot recognition.

However, a key challenge remains: how to effectively represent event streams, which consist of events characterized by timestamps, pixel positions, and polarity. Obviously, directly feeding this raw event data into an LLM is infeasible (see Section~\ref{results}). To address this problem, we explored two representative methods for representing event signals: the first involves integrating all events based on their positions on a frame, resulting in what is termed an ``event frame''. The second method leverages event-to-video techniques, such as E2VID \cite{b10} and E2HQV \cite{b11}, to reconstruct the event signals into human-understandable natural images, referred to as ``reconstructed frame''.

Our key contributions and findings can be summarized as follows:
\begin{itemize}
    \item To the best of our knowledge, our work is the first comprehensive exploration and evaluation of the pure zero-shot recognition capabilities of LLMs, such as GPT-4o, on the N-ImageNet \cite{b12} dataset, exceeding the accuracy of state-of-the-art methods by five orders of magnitude;
    
    \item We systematically evaluate the impact of different representations of event data inputs on the recognition ability of LLMs. The results indicate that using model-reconstructed frames typically improves recognition accuracy;
    
    \item This work provides guidance for potential research directions in applying LLMs to event-based visual content and can serve as a benchmark for future zero-shot recognition studies based on event camera data.
\end{itemize}

\section{Related Work}
\noindent\textbf{Exploration of LLMs.}
Recent studies have conducted in-depth explorations of LLMs' capabilities across various tasks, including zero-shot recognition \cite{b13}, the 2023 BioASQ challenge \cite{b14}, zero-shot classification of point cloud data \cite{b15}, emotion recognition \cite{b16}, and visual recognition and localization \cite{b17}. Although substantial research has explored various visual tasks, there is still a lack of quantitative analysis on the performance of LLMs in event-based pure zero-shot recognition. Furthermore, to the best of our knowledge, only a limited number of studies have evaluated the effectiveness of open-source LLMs in visual tasks.
\vspace{3pt}

\noindent\textbf{Event-based zero-shot visual recognition.}
The web-scale image-text pre-training model CLIP \cite{b8} has bridged visual and textual domains, inspiring extensions into event-based scene understanding. EventCLIP \cite{b7} adapted CLIP for event-based recognition by converting event data into 2D representations and refining feature alignment. ECLIP \cite{b4} and EventBind \cite{b5} further improved this approach by incorporating an event encoder and a Hierarchical Triple Contrastive Alignment (HTCA) module. However, these methods heavily rely on extensive training and CLIP's limitations, requiring specialized event encoders.

\section{Datasets and Task Setups}
We select three representative event datasets for recognition experiments. To ensure a fair comparison, these datasets undergo the same pre-processing steps before being fed into the LLMs. After obtaining the model's outputs, we apply post-processing to the results.
\vspace{3pt}

\noindent \textbf{Dataset.} We use three experimental datasets: N-ImageNet \cite{b12}, N-Caltech101 \cite{b18}, and N-MNIST \cite{b19}. N-ImageNet is derived from ImageNet-1K \cite{b20}, where RGB images are shown on a monitor and captured by a moving event camera, resulting in 1,781,167 event streams at 480 × 640 resolution across 1,000 object classes. N-Caltech101 consists of 8,246 event streams, each 300 ms long, recorded by an event camera capturing images from Caltech101 \cite{b19} displayed on an LCD monitor, covering 101 object categories. N-MNIST includes 70,000 event streams, each representing one of the 10 handwritten digits from the MNIST \cite{b21} dataset.
\vspace{3pt}

\begin{figure}[t]
  \centering
  \includegraphics[width=0.45\textwidth]{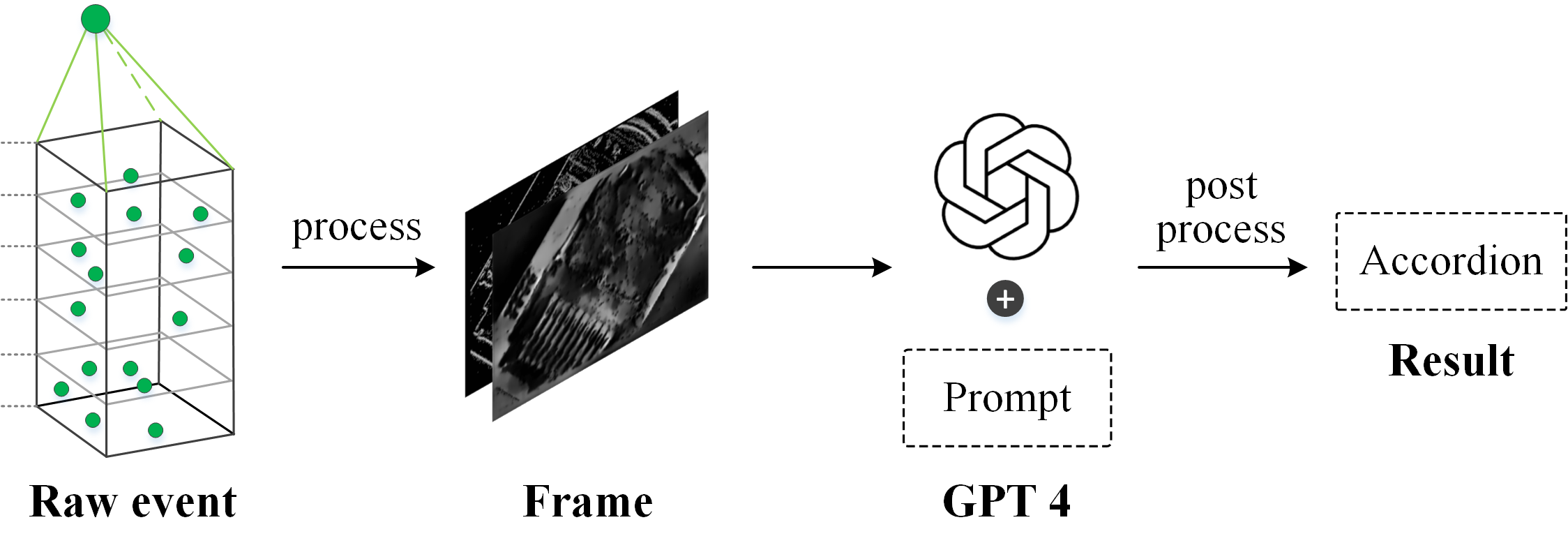}
  \caption{\textbf{Workflow for event-based visual content recognition using LLMs.} The process starts by converting raw event stream into event frames, or alternatively, into reconstructed frames using \cite{b7} and \cite{b8}. These converted frames, along with a prompt, are then input into the LLM for recognition. Once the model returns its results, post-processing is performed to obtain the final outcome.}
  \label{fig:flow}
\end{figure}

\noindent\textbf{Dataset processing.} As shown in Figure \ref{fig:flow}, we convert the raw event stream from three experimental datasets into two formats: event frames and reconstructed frames. First, we transform the event streams into event frames. Specifically, we process the raw event data by projecting the voxel grid onto a 2D histogram, applying logarithmic scaling for normalization, and mapping the result to an 8-bit event frame to represent event frequency. Next, we use E2VID \cite{b10} and E2HQV \cite{b11} to generate reconstructed frames from the event signals. For the N-ImageNet and N-Caltech101 datasets, we segment the event streams into 0.05-second intervals and select the middle interval of each stream to generate two different representations of frames for the test. Since some event streams in the N-ImageNet dataset are shorter than 0.05 seconds, we convert all of those streams into a single event frame. Given the typically sparse event streams in the N-MNIST dataset, we combine all event data into a single frame. To ensure compatibility with the input format required by E2HQV and to enable fair comparison, we resize all event data from the three datasets to a resolution of 180 × 240 before converting them into frames. After converting the event streams into frames, we upload them to the cloud. During testing, the models access the frames via URLs, and all four models support reading frames from URLs.

\subsection{Task Setups}
We evaluate the pure zero-shot recognition capabilities of several LLMs for event-based visual content.
\vspace{3pt}

\noindent\textbf{Task.} As illustrated in Figure \ref{fig:flow}, our task is to present event frames or reconstructed frames generated from raw event streams to a LLM and determine whether it can accurately identify the category to which each frame belongs. The model is given its task along with multiple options (e.g., prompts), but only one option is correct. The number of options corresponds to the total number of categories in each dataset: 1,000 options for N-ImageNet, 101 options for N-Caltech101, and 10 options for N-MNIST. 
\vspace{3pt}

\noindent\textbf{Evaluation metrics.}
The metric used to evaluate the model’s performance is accuracy, which is the proportion of correctly classified instances out of the total number of instances presented. This metric provides a straightforward assessment of the model’s overall effectiveness in the classification task.
\vspace{3pt}

\noindent\textbf{Post-processing.}
Since different models may produce varying responses, we need to apply post-processing to match the correct answer, as shown in Figure \ref{fig:flow}. This includes removing all characters except for the answer, converting all letters to lowercase, and performing character matching. For GPT-4o, due to its superior performance, we can add constrained prompts to ensure it returns the desired response format, eliminating the need for post-processing.

\section{Experiments}
\noindent\textbf{Models}
\vspace{3pt}

\noindent We explore four LLMs, including two of the most powerful models from OpenAI and two representative open-source models that can process visual content. All dataset representations are evaluated in the same sequence across these models.
\vspace{3pt}

\noindent\textbf{GPT-4o, GPT-4turbo.}
We utilize two OpenAI models as pure zero-shot models. We provide a task description to each model, and the model returns the corresponding answer. For GPT-4o, we use the latest version, chatgpt-4o-latest \cite{b22}, which was released on August 15, 2024. For GPT-4-turbo, we use the most recent version, gpt-4-turbo-2024-04-09 \cite{b22}, which the GPT-4-turbo model points to.
\vspace{3pt}

\noindent\textbf{LLaVA.}
We use LLaVA as one of our pure zero-shot models. Similarly, we provide the model with a task description, and it returns the corresponding answer. Specifically, we use the LLaVA-v1.5-7b version \cite{b23}.
\vspace{3pt}

\noindent\textbf{MiniGPT-4-v2.}
We also use MiniGPT-4-v2 as one of our pure zero-shot models. Similarly, we provide it with a task description, and it returns the corresponding answer. We specifically use the MiniGPT-4-v2 \cite{b24} model based on LLaMA2 Chat 7B \cite{b25}, with stage 3: Multi-modal instruction tuning \cite{b24}, as the pretrained weights.

\vspace{3pt}

\noindent\textbf{Hardware and software details}
\vspace{3pt}

\noindent We run LLaVA and MiniGPT-4-v2 on an RTX 3090. Since GPT models only require API calls, we set up the environment and call the GPT API on an RTX 3050.

\begin{table}[h]
\centering
\caption{Comparison of Classification Accuracy Across N-ImageNet (N-I), N-Caltech101 (N-C), and N-MNIST (N-M) Datasets. Due to the prompt length limitations of LLaVA and MiniGPT-4-v2, we are unable to evaluate their recognition performance on the N-ImageNet dataset. \textbf{Bold values indicate the best performance for each dataset.}}
\begin{tabular}{ccccc}
\hline
\multirow{2}{*}{Method} & \multirow{2}{*}{Input} & \multicolumn{3}{c}{Accuracy (\%)} \\ \cline{3-5} 
                        &                             & N-I  & N-C  & N-M  \\ \hline
ECLIP                   & Event frame                       & 8.72        & 53.88         & 14.56      \\
EventCLIP               & Event stream                       & 4.30        & 49.95         & 11.87      \\ 
EventBind               & Event frame                      & 3.05        & 67.58         & 12.89      \\  \hline 
                        & Event frame                         & -           & 23.10         & 15.29      \\ 
MiniGPT-4-v2\textsuperscript{*}              & E2VID                       & -           & 30.20         & 10.70      \\ 
                        & E2HQV                       & -           & 30.71         & 11.62      \\ \hline
                        & Event frame                         & -           & 64.72         & 56.27      \\ 
LLaVA\textsuperscript{*}                   & E2VID                       & -           & 67.26         & 19.57          \\ 
                        & E2HQV                       & -           & 67.51         & 22.94          \\ \hline
                        & Event frame                         & 17.23       & 72.08         & 85.95          \\ 
GPT-4turbo                  & E2VID                       & 11.61       & 55.08         & 39.14          \\ 
                        & E2HQV                       & 10.70       & 55.58         & 39.41          \\  \hline
                        & Event frame                         & \textbf{45.55}           & 89.37         & \textbf{91.86}      \\ 
GPT-4o                  & E2VID                       & 32.75      & \textbf{93.90}         & 49.79      \\ 
                        & E2HQV                       & 30.05       & 93.23         & 51.68      \\ \hline
\end{tabular}
\label{tab:comparison_accuracy}
\footnotesize \textsuperscript{*}Open-source model
\end{table}

\subsection{Results}
\label{results}
Table \ref{tab:comparison_accuracy} presents the performance of LLM-based pure zero-shot method we evaluated and compares it with other state-of-the-art zero-shot event-based object recognition methods. Due to the limited prompt input length in LLaVA and MiniGPT-4-v2, we are unable to evaluate their performance on the N-ImageNet dataset.
Our evaluations address the following seven questions.
\vspace{3pt}

\noindent\textbf{Q1: Can LLMs understand and recognize raw event streams?} \textit{Test: GPT-4o.} Answer: \textbf{No.} When given raw event streams and asked to identify their corresponding category, GPT-4o responds by stating that these are raw event streams and that it cannot directly analyze them. These models are designed to handle natural language, code, or structured data, while event camera outputs low-level, asynchronous data (e.g., pixel position, timestamp and polarity), representing spatiotemporal information. Thus, LLMs cannot directly understand or recognize raw event streams.

\vspace{3pt}
\noindent\textbf{Q2: Can using more suitable prompts improve the model's effectiveness?} \textit{Test: Original prompt vs. Improved prompt.} Answer: \textbf{Yes.} When we simply instruct the model to classify the given frames, its accuracy is relatively low. However, when we first explain to the model what event streams are and specify the representation of the frames provided to it before asking it to classify, the model's recognition accuracy improves by approximately 2\%.
\vspace{3pt}

\begin{figure}[t]
  \centering
  \includegraphics[width=0.48\textwidth]{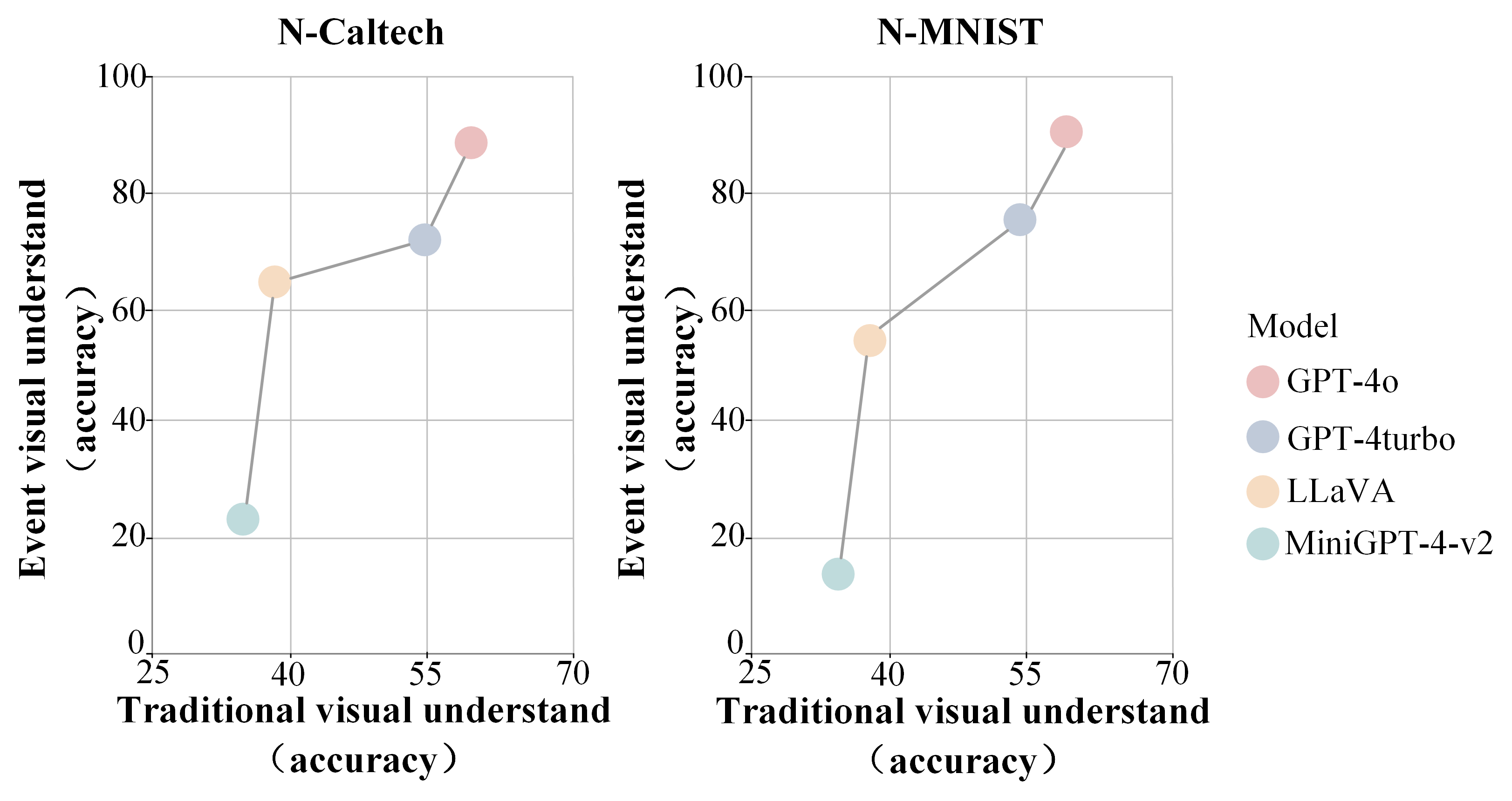}
  \caption{\textbf{Comparison of LLMs' understanding of event-based visual content and traditional visual content.} A contrast between the event-based visual content recognition abilities of the evaluated LLMs and the traditional visual content perception abilities of the models assessed in \cite{b9}.}
  \label{fig:event_vs_traditional}
\end{figure}

\noindent\textbf{Q3: Can using the event2video models to generate reconstructed frames improve recognition accuracy compared to event frames?} \textit{Test: Event frames vs. E2VID vs. E2HQV.} Answer: \textbf{Yes, but it depends on the quality of the event dataset.} For the N-Caltech101 dataset, which features relatively high resolution, downsampling to match E2HQV's input dimensions and generating reconstructed frames effectively maintains the integrity of the original event stream without introducing significant noise. As indicated in Table \ref{tab:comparison_accuracy}, when utilizing LLaVA for N-Caltech101, the accuracy of E2VID and E2HQV reconstructed frames exceeds that of event frames by $2.54\%$ and $2.79\%$, respectively. When tested with GPT-4o, this improvement is even more pronounced, with increases of $4.53\%$ and $3.86\%$. In contrast, GPT-4turbo demonstrates better performance with event frames, likely due to the higher noise present in the reconstructed frames. Given GPT-4turbo's limitation to text processing, it is less proficient in multimodal classification tasks, whereas GPT-4o, with its multimodal capabilities, delivers superior cross-modal performance.

Due to the low resolution of N-MNIST (35 × 35), retaining the original resolution and padding with a blank background for E2HQV processing introduces significant noise in areas without events, leading to poor reconstruction quality. Consequently, this approach degrades rather than improves recognition accuracy. As shown in Table \ref{tab:comparison_accuracy}, when tested with GPT-4turbo, the accuracy of event frames is more than double that of the reconstructed frames.

For the N-ImageNet dataset, the accuracy of event frames is higher than that of reconstructed frames. This is because, as previously mentioned, many event streams in N-ImageNet are shorter than 0.05 seconds. When we convert all these event streams into reconstructed frames, the excessive number of events can negatively impact the quality of the reconstruction.
\vspace{3pt}

\noindent\textbf{Q4: Does GPT-4o outperform GPT-4turbo?} \textit{Test: GPT-4o vs. GPT-4turbo.} Answer: \textbf{Yes.} The experimental results clearly show that GPT-4o outperforms GPT-4turbo in recognition across all three datasets. Notably, its performance significantly surpasses that of the other two open-source models, highlighting GPT-4o's exceptional ability to understand event-based visual content.
\vspace{3pt}

\noindent\textbf{Q5: Does stronger traditional visual understanding in LLMs typically indicate better event-based visual content understanding?} \textit{Event-based recognition vs. Traditional visual perception.} Answer: \textbf{Yes.} Figure \ref{fig:event_vs_traditional} illustrates the event-based visual content recognition capabilities of each model and compares these results with traditional visual content perception as reported in \cite{b9}. It is evident that the stronger a model's understanding of traditional visual content, the better its ability to comprehend event-based visual content, indicating a positive relationship. Notably, once a model's traditional visual comprehension reaches a certain threshold, it exhibits emergent abilities, leading to a significant improvement in its event-based visual understanding. Specifically, as shown in Figure~\ref{fig:event_vs_traditional}, when comparing LLaVA to MiniGPT-4-v2, although LLaVA's traditional visual understanding improves by less than 10\%, its event-based visual comprehension increases by approximately 40\%. Therefore, we hypothesize that as LLMs continue to advance, their ability to understand event-based visual content will also improve accordingly.
\vspace{3pt}

\section{Conclusion}

In this paper, we evaluate a pure zero-shot event-based recognition method using LLMs and demonstrate its superiority over existing state-of-the-art methods in zero-shot tasks. We also explore how different event representations impact LLM recognition capabilities, finding that converting event streams into reconstructed frames can improve accuracy. These findings highlight the potential of LLMs in event-based visual content understanding, underscore their promising applications in this domain, and provide a foundation for future research in event-based visual content recognition. In the future, we will continue to explore advanced methods to enhance the accuracy of pure zero-shot event-based recognition.




\end{document}